\documentclass{article}
\usepackage{kr}

\usepackage{times}

\usepackage{soul}
\usepackage{url}
\usepackage[hidelinks]{hyperref}
\usepackage[utf8]{inputenc}
\usepackage[small]{caption}
\usepackage{graphicx}
\usepackage{amsmath}
\usepackage{amsthm}
\usepackage{booktabs}
\usepackage{algorithm}
\usepackage{algorithmic}
\usepackage{todonotes}
\usepackage{amssymb}
\usepackage{natbib}

\usepackage[inline]{enumitem}

\newtheorem{example}{Example}

\newcommand{\change}[1]{\textcolor{black}{#1}}
{}

\usepackage{todonotes}

\title{Contestable AI needs Computational Argumentation}

\author{%
Francesco Leofante\and
Hamed 	Ayoobi\and
Adam 	Dejl\and
Gabriel 	Freedman\and \\
Deniz 	Gorur\and
Junqi 	Jiang\and
Guilherme 	Paulino-Passos\and
Antonio 	Rago\and \\
Anna 	Rapberger\and 
Fabrizio 	Russo\and
Xiang 	Yin\and
Dekai 	Zhang\And
Francesca Toni\\
\affiliations
Computational Logic and Argumentation Group, \\
Department of Computing, Imperial College London, UK\\
\emails
\{f.leofante, h.ayoobi, adam.dejl18, g.freedman22, d.gorur22,
	junqi.jiang20,
	gppassos,
	a.rago,
	a.rapberger,
	fabrizio, x.yin20,
	dekai.zhang19, f.toni\}@imperial.ac.uk
}

\begin{document}

\maketitle

\begin{abstract}
AI has become pervasive in recent years, but state-of-the-art approaches 
predominantly neglect the need for AI systems to be \emph{contestable}.  
Instead, contestability is advocated by AI guidelines (e.g. by the OECD) and 
regulation of automated decision-making (e.g. GDPR). 
In this position paper we explore how contestability can be achieved computationally in and for AI. 
We argue that \emph{contestable AI} requires dynamic (human-machine and/or machine-machine) explainability and decision-making processes, whereby machines can 
\begin{enumerate*}
    \item interact with humans and/or other machines 
to progressively explain their outputs and/or  their reasoning as well as assess grounds for contestation provided by these humans and/or other machines, and
    \item revise their decision-making processes to redress any issues successfully raised during contestation.
\end{enumerate*} Given that much of the current AI landscape is tailored to static AIs, the need to accommodate contestability will require a radical rethinking, that, we argue,  computational argumentation is ideally suited to support.
\end{abstract}

%\keywords{Contestable AI, Explainable AI, Interactivity}

%%%%%%%%%%%%%%%%%%%%%%%%%%%%%%%%%%%%%%%%%%%%%%%%%%%%%%%%%%%%%%%%%%%%%%%%

%%%%%%%%%%%%%%%%%%%%%%%%%%%%%%%%%%%%%%%%%%%%%%%%%%%%%%%%%%%%%%%%%%%%%%%%

\section{Introduction}

AI has become pervasive in recent years, with applications ranging from autonomous driving~\citep{muhammad2020deep} to finance~\citep{cao2022ai} and healthcare~\citep{shaheen2021applications}. 
% \todo{add refs, examples}
Current state-of-the-art AI systems focus on algorithmic solutions, often built from data, generating outputs in the forms of predictions, recommendations and/or decisions and, in some cases, explanations  thereof. These existing solutions, though, mostly neglect  the need for these AI systems to be \emph{contestable}.      
Instead,  the need to accommodate contestability is a crucially important problem if AI systems are to be deployed in and benefit society. Indeed, 
contestability is prominently advocated in some frameworks for AI ethics and in regulations. For example, the Organisation for Economic Co-operation and Development (OECD) states that information should be provided ``to enable those adversely affected by an AI system to challenge its outcome based on plain and easy-to-understand information on the factors, and the logic that served as the basis for the prediction, recommendation or decision'' (Principle 1.3\footnote{\url{https://oecd.ai/en/dashboards/ai-principles/P7}}). Moreover,  
 the ACM Global Technology Policy Council for Responsible Algorithmic Systems advocates `Contestability and Auditability' as one of nine principles.\footnote{\url{https://www.acm.org/media-center/2022/october/tpc-statement-responsible-algorithmic-systems}}
Contestability is required by law in some jurisdictions, e.g. 
 GDPR, article 22(3),\footnote{\url{https://gdpr-text.com/read/article-22/}}  states that the data subject’s rights to be safeguarded shall include ``at least the right […] to contest the decision''. Finally, contestability and redress are also identified as key principles underpinning the UK AI regulation framework\footnote{https://www.gov.uk/government/publications/ai-regulation-a-pro-innovation-approach/white-paper}\change{, whereby ``Guidance should clarify existing ‘formal’ routes of redress offered by regulators in certain scenarios". This is a major departure point from ``classical" requirements for explainable systems, which typically do not require correcting the AI when faulty behaviours are exposed.}

 Contestability is seen by some~\citep{Hicks22} as a means to facilitate accountability, e.g. to prevent the use of Uber's algorithm  for banning drivers because it is not taking into account bias of customers when giving bad reviews. A handful of 
recent works (overviewed in Section~\ref{sec:related}) bring contestability to the attention of AI researchers, practitioners and users, while providing insights into its possible interpretations in practice.
Moreover,  there is evidence, in the form of user studies, that  
%varying levels of explanations, human oversight, and contestability for high- and low-stakes algorithmic loan approval scenarios affect users’ informational, procedural, and overall fairness perceptions
contestability %, alongside explainability, 
can affect users' perception of AI fairness~
\citep{Yurrita23}. However, formal/computational forms of contestability are mostly lacking in the literature, with some exceptions, in the form of 
 \begin{enumerate*}
     \item indications that, in the case of machine learning, methods generating counterfactual explanations in terms of ``actionable changes that an individual can make to flip the prediction of the classifier'' may be interpreted as offering (limited) contestability~\citep{Alfano20} and
     \item \citep{Fab-ECAI23}, offering contestability in the very specific setting of causal discovery.
 \end{enumerate*}
 % 
 %Furthermore, existing interpretations of contestability  of AI systems \todo{predominantly? exclusively?} focus on humans contesting machines, thus ignoring the possibility that it may be beneficial for machines to contest humans or other machines.  
 Finally, existing works mostly see contestability as a post-hoc process, after predictions/recommendations/decisions have been computed.

In this position paper (specifically in Section~\ref{sec:view}) we consider what it means for (any)
%\emph{contestable AI} may be achievable REDO??? in \emph{multi-agent systems}, where agents may be humans or 
\emph{algorithmic decision systems (ADS)}~\citep{HeninM22_justifiability} to be contestable and contested (by humans or other ADSs). 
%Specifically,  humans may range from developers of ADSs, to regulators, expert users, and lay users, and  ADSs could be machine learning models for classification, rule-based systems for inference, and so on. 
Our take-away message is that 
 formalising/realising computationally contestable AI  %in multi-agent systems 
%is a promising direction of research as soon as ADSs can exchange explanations while interacting.
requires: \emph{explanations} (by the ADSs which humans/other ADSs may want to contest); \emph{grounds
} for contestation (by the contesters); ability (by the contested) to \emph{redress} any issues successfully raised during contestation; and ability to \emph{interact} (by both contested and contester). 
We then 
advocate computational argumentation (CA) as being ideally suited to support contestability, providing evidence from the literature on CA~(Section~\ref{sec:CA}).
%a concept of \emph{contestable AI}  whereby two or more agents can engage in \emph{two-ways interactions} in order to challenge, correct, or improve (automated  or human) decisions \emph{as or after} they are made, supported by \emph{explanations}. 
We finally conclude (Section~\ref{sec:conc}) with some pointers for future directions in what we believe is a very promising direction of research for the KR community.

\section{AI Contestability %: a bird's eye view
in the literature}
\label{sec:related}

% \todo[inline]{we do not have space for a table. I would write a summary in text. We need to find some dimensions. May be following the columns: all assume that the contestation is done by humans of one type or another (e.g. psyco for X etc), that the contested is an ADS (of which type?) or its outputs, given some inouts, or data.....}

Here, we summarise recent works %in the literature 
focusing on advocating or proposing forms of contestability in AI.\footnote{We have identified these works by searching DBLP with `Contestable' and `Contestability', restricting attention to papers on AI from 2014 onwards published in peer-reviewed venues, while also considering additional references therein.} We %summarise 
focus on the main contestability dimensions that emerge in this literature, as well as providing pointers to other (less) related research.

% In our initial discussion we focus on three key questions:
% \begin{itemize}
%     \item What is being contested?
%     \item Who is doing the contesting? 
%     \item How may contestation be realised (in terms of the method used to support it)?
% \end{itemize}
% before diving deeper into selected aspects of contestability discussed in prior works.

\textbf{Contested entities.} Prior work has %predominantly 
focused on contesting various aspects of an ADS, including its general design~\citep{Almada19,alfrink2022contestable,yurrita-process-centric-explanations}, %the used
training data~\citep{kluttz2022shaping}, training procedure~\citep{Almada19}, its inputs \citep{alfrink2022contestable} and outputs~\citep{HirschMNIA17,Ploug_20,TubellaTDM20,LyonsVM21,HeninM22_justifiability,Hicks22} as well as the %system 
ADS as a whole~\citep{LyonsVM21,VaccaroXHK21_Moderation,HeninM22_justifiability,Hicks22,cameracars23,Fab-ECAI23}.  %In contrast, 
Note that the possibility of  humans being contested by an ADS (e.g., if their assumptions about the subject domain are flawed) %has 
is not typically considered.

\textbf{Contesting %agents
entities.} All %prior 
approaches %that 
we identified %during our review focused 
focus on humans as the contesting %agents
entities. However, there are considerable differences between the categories and assumed skill sets of people engaged in %a
contestation. Several argue that decision subjects should be empowered to directly contest ADS decisions affecting them, potentially without possessing
detailed knowledge of the subject domain \citep{Almada19,Ploug_20,LyonsVM21,VaccaroXHK21_Moderation,HeninM22_justifiability,Hicks22}. Some works 
suggest that individual decision subjects may not always be able to effectively contest on their own, and propose contestation via third-party representatives
\citep{alfrink2022contestable} or as part of a group in a ``class action" \citep{LyonsVM21}. Others also consider contestation by professionals, subject matter experts and regulators \citep{TubellaTDM20,HeninM22_justifiability,kluttz2022shaping,cameracars23,Fab-ECAI23}.
Note that the possibility of   ADSs being the contesting entities is not typically 
 considered.

\textbf{Contestation methods.} Several methods have been proposed for facilitating contestation. In the realm of non-technical solutions, participatory design has been advocated as a way to involve various stakeholders in ADS development, enabling advance mitigation of possible issues and risks associated with the use of the ADS \citep{Almada19,VaccaroXHK21_Moderation,cameracars23}. The framework proposed in \citep{alfrink2022contestable} suggests following a set of development practices over the ADS lifecycle, including incorporating ex-ante safeguards, gathering end-user feedback, implementing quality assurance, mitigating possible risks and allowing for third-party oversight. More technical approaches envision usage of operation logs \citep{Hicks22} or model explanations \citep{Almada19,Ploug_20,kluttz2022shaping,Fab-ECAI23,yurrita-process-centric-explanations}. We discuss these and their use of explanations in greater detail %in the following section
next.

% Table~\ref{summary-table} provides a summary of recent works in the literature focusing on advocating or proposing forms of contestability in AI.\footnote{We have identified these works by searching DBLP with `Contestable' and `Contestability', restricting attention to papers on AI from 2014 onwards published in peer-reviewed venues, while also considering some additional references therein. } 

% We list the identified papers in chronological order, focusing on how they address the following questions:
% \begin{itemize}
%     \item what is being contested?
%     \item who is doing the contesting? 
%     \item %which methods are proposed for supporting this contesting
%     how may contestation be realised (in terms of the format the contestation may take and methods to support it)?
% \end{itemize}
%

%\todo[inline]{the discussions below should be mostly focused on papers in the table - with some extra ones possibly: to be done}
%\todo[inline]{double-check what is being contested in each paper and link it to our abstract model (in later sections), especially focus on what it means to contest the input and the output (also contesting the domain?)}

\textbf{The role of explanations% and contestability
.} That explanations are needed to support contestability is acknowledged by some (e.g. by \cite{Almada19,LyonsVM21,alfrink2022contestable,wachter}), but the problem of using explanations to support contestability in practice has received little attention in explainable AI (XAI) (or AI for that matter) to date. 
   A notable exception is~\citep{Fab-ECAI23}, where causal discovery with neural networks guides human feedback for contesting the discovered causal relations, which can be seen as a form of global explanation.  
   Also, some current XAI methods can be seen as offering ground for some limited contestability when they suggest actionable recourse, as is the case for counterfactual explanations in terms of ``actionable changes that an individual can make to flip the prediction of the classifier”~\citep{Alfano20}. However, these works  disregard that algorithmic decisions may actually be incorrect, e.g. because they are based on incorrect or incomplete data.  Moreover, these methods are one-shot, providing no opportunities for follow-up inquiry, and %zero-knowledge
   shallow, revealing no information about the steps or logic that led to the explained output, and thus offer little ground for contestability. 
   Some other XAI methods provide information on the AIs’ deliberation and insecurities~\citep{wang2019deliberative}; however, they do not support any form of contestability. Overall, explanations, as understood in state-of-the-art XAI, are seen as inadequate to support contestability~\citep{HeninM22_justifiability}. Specifically, \cite{yurrita-process-centric-explanations}
   advocate a generic notion of explanations that  capture the rationales behind the development and deployment of the ADS (referred to therein as ``process-centric explanations'').

   % Added as yurrita-process-centric-explanations, but seems relatively limited
%\todo[inline]{May consider adding cite{yurrita-process-centric-explanations} here, although these are not directly connected to XAI in technical sense — their process-centric explanations mainly focus on the process for the AI development, similarly to datasheets for datasets and model cards}

\textbf{Post-hoc and ex-ante contestability.} Some %works 
take the view that contestability needs to be supported by regulatory frameworks, involving in particular policy modelling and normative reasoning \citep{TubellaTDM20}. These approaches see contestability as post-hoc processes, detached from explainability, “to review algorithmic decisions”~\citep{LyonsVM21}. Ex-ante contestability is envisaged by some as a design principle \citep{HirschMNIA17,alfrink2022contestable}, whereby systems are designed to enable users to interactively contest these systems and interaction between humans and systems is needed for ``critique and correction” \citep{LyonsVM21}, but no technical solutions exist on how to support it. % Further, all existing discussions envisage contestability by humans towards ADSs, while ignoring  that  it may be beneficial for ADSs to be able to contest too, e.g. to critique humans or other ADSs’ biases. 
 
\textbf{Contestability as an interaction process.}
\cite{kluttz2022shaping} see contestability as an interaction process with humans, at development and deployment times, to allow humans, rather than just data, to train systems. Several other works (e.g.~\citep{HirschMNIA17,LyonsVM21}) envisage post-hoc contestability as an interaction process, again between humans and ADSs alone.

\textbf{%Corrigibility and the AI alignment problem
Other related work.}
%Closely related to contestability %and redress 
%are 
Works on \emph{corrigibility}, such as~\citep{DBLP:conf/aaai/SoaresFAY15,Carey,DBLP:journals/aim/RussellDT15} % (see \cite{Carey} for a recent review of %literature on corrigibility under a unifying formal framework)
%. These works 
focus on  human overseers %(users, developers, regulators) 
providing feedback/instructions to ADSs
so that they align 
with the intentions/values of their user(s)  \citep{DBLP:journals/mima/Gabriel20,DBLP:conf/nips/Hadfield-Menell16}. %The AI alignment problem
While corrigibility can be viewed as enabling human users to contest the decisions of ADSs% with explicit corrective feedback and instruction
, 
%they are also limited to contestability of ADSs by humans.
%Moreover, corrigibility  %important when the ADSs are AI systems 
it is narrowly
focused on ADSs which optimise for reward %, as they may often 
when they have %default 
incentives to manipulate users \citep{DBLP:conf/atal/WardTB22}, ignore instructions \citep{DBLP:conf/nips/Hadfield-Menell16}, and disempower humans/other ADSs \citep{DBLP:conf/nips/TurnerSSCT21}.  

\section{An abstract view of AI contestability}
\label{sec:view}

\begin{figure}
    \centering
    \includegraphics[width=0.35\textwidth]{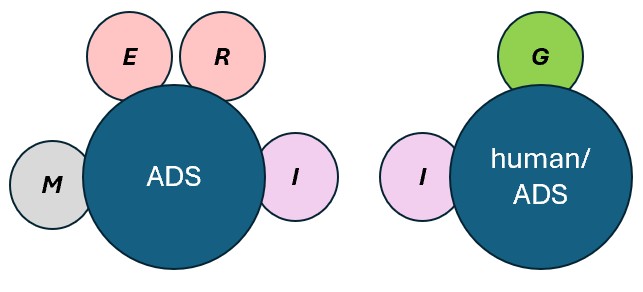}
    \caption{%ADSs for
    An abstract view of AI contestability: the contested ADS (left) is equipped with a model ($M$), an \emph{explanation} method ($E$), and a \emph{redress} method ($R$); the contester (right) is a human or an ADS equipped with a \emph{ground generator} for contestations ($G$); both contested ADS and contester are able to interact ($I$)% with one another
    .}
    \label{fig:overview}
\end{figure}

In a nutshell, towards supporting contestability, we see ADSs and humans as in Figure~\ref{fig:overview}, assuming that the contested entity is an ADS and the contester is either a human or another ADS.\footnote{For simplicity, we assume that the contested ADS is equipped with a single $M$ and accompanying $E$ and $R$, but in general the same ADS may be equipped with several. Similarly, we assume that the %(human/ADS) 
contester is equipped with a single $G$, but in general 
the same contester may be equipped with several (one per contested model% being contested
). Further,  the same ADS could %in principle 
be both contested and contester%, for different models
. }  

Let $M:I \rightarrow O$ be a \emph{model} computing outputs in $O$ for inputs in $I$.
$M$ could be. e.g., a machine learning model, an expert system, or a combination of the two%, amongst several
. $M$ is part of the (contested) ADS, and the outputs of $M$ are the ADS's decisions.
For this ADS to be contestable (by a human or another ADS), it needs
to be able to process \emph{contestations}, which may be in reference to one of the following  settings:\footnote{We ignore here the trivial setting where contestations are directed at specific inputs, as these do not require altering $M$.}

\textbf{(A)} output $y=M(x)$ for some specific input $x$, or
%\todo{Also accommodates Jay's example}

\textbf{(B)} how $M$ determines $y=M(x)$ from input $x$, or

\textbf{(C)} the full model $M$.
\\
The first setting (A)  amounts to contesting %external 
the model behaviour without referring to its internal ``reasoning'' process. The second setting (B) amounts to contesting (part of) the reasoning process of the model for the particular input. The third setting (C) amounts to contesting the model in general, e.g. its ``philosophy'', such as the underpinning input and output distributions and/or that it is  biased.
%\todo[inline]{Possible examples for case 3: fixing general bias, LLM hallucinations, type of model (black-box vs transparent)}

\iffalse 
\begin{example}
As a simple illustration, consider $M$ given by a (transparent) set of rules for binary classification ($c_1,c_2$) with binary features ($f_1, f_2$):
\begin{eqnarray*}
&& \forall X [f_1(X) \wedge \neg f_2(X) \rightarrow p(X)] \label{eq:rule1}\\
&& \forall X [p(X) \rightarrow c_1(X)]\\
&& \forall X [\neg p(X) \rightarrow c_2(X)] 
\end{eqnarray*}
Then, for input $x$ characterised by features $f_1$ and $\neg f_2$, classified as $c_1$ by $M$, in the first setting a contestation may amount to objecting to the classification $c_1$, whereas in the second setting it may amount to objecting to the instance of the top rule for $x$ (i.e. $f_1(x) \wedge \neg f_2(x) \rightarrow p(x)$).
%In both cases the contestation may also refer to the input, e.g. that the model wrongly sees $x$ as characterised by $\neg f2$.
The top rule could also be ground for contestation in the third setting, but without any reference to the specific input $x$.
\end{example}
\todo{have a concrete example instead?}
\fi

\begin{example}
\label{ex:loan}
%\todo{this example can be changed - but I would stick to a set of rules. If changed ripple effect though, as used in several places...} 
As a simple illustration, consider $M$ given by a (transparent) set of rules for binary classification amounting to awarding a loan (class $\texttt{loan}$), or not (class $\neg \texttt{loan}$), with binary features %$gender$, 
$employed$ ($\texttt{emp}$ in short) and $career\_breaks$ ($\texttt{breaks}$ in short):\footnote{These rules may correspond to the behaviour of a machine learning classifier%, e.g a decision tree or multi-layer perceptron
, where $\texttt{continuous\_emp}$ is a ``latent feature'' of the model. We use them informally here (e.g. they may be implications in classical logic or logic programming rules, with $\neg$ as negation as failure).}
\begin{eqnarray*}
&&\hspace{-0.65cm}\forall X [\texttt{emp}(X) \wedge \neg \texttt{breaks}(X) \rightarrow \texttt{continuous\_emp}(X)] \label{eq:rule1}\\
&&\hspace{-0.65cm}\forall X [\texttt{continuous\_emp}(X) \rightarrow \texttt{loan}(X)]\\
&&\hspace{-0.65cm}\forall X [\neg \texttt{continuous\_emp}(X) \rightarrow \neg \texttt{loan}(X)]     
\end{eqnarray*}
For an input (loan applicant) $x$ characterised by features $\texttt{emp}$ and $\texttt{breaks}$,
classified as $\neg \texttt{loan}$ by $M$, in the first setting a contestation may amount to objecting to the classification $\neg \texttt{loan}$, whereas in the second setting it may amount to objecting to the instances for $X=x$  of the top and bottom rules (i.e. $\texttt{emp}(x) \wedge \neg \texttt{breaks}(x) \rightarrow \texttt{continuous\_emp}(x)$
and $\neg \texttt{continuous\_emp}(x) \rightarrow \neg \texttt{loan}(x)$).
%In both cases the contestation may also refer to the input, e.g. that the model wrongly sees $x$ as characterised by $\neg f2$.
In the third setting, the middle rule could  be the object of contestation, without reference to any specific input.
\end{example}

\textbf{Explanations for contestability.} In line with some of the literature,
we envisage that contestability needs to be supported by
explanations for $M$ and/or its outputs for specific inputs. We thus assume that the ADS, in addition to $M$, is equipped with an \emph{explanation method} $E$. Differently from \cite{yurrita-process-centric-explanations}, we believe that technical notions  of explanations from the literature in XAI may already prove useful. These explanations may be \emph{local} (i.e. relating to a specific input, output pair)  or \emph{global} 
(i.e. relating to all input, output pairs). Different forms of contestability  may require and/or benefit from different forms of explanations and thus explanation methods, including as follows:
\begin{enumerate*}
    % \item $E$ is a \emph{local explanation} method; this may amount to
    % \begin{itemize} 
    \item %an  \emph{input-output explanation} method, e.g. 
    a \emph{feature-attribution explanation} method, e.g. as provided by model-agnostic LIME \citep{lime} or SHAP \citep{shap}  or  by model-specific (tailored to neural classifiers) LRP \citep{lrp} or DeepLIFT \citep{deeplift};
    \item an \emph{abductive explanation} method, e.g. as in \citep{IgnatievNM19};
    \item a \emph{counterfactual explanation} method   e.g. as first envisaged in \citep{wachter,TolomeiSHL17}; 
    \item 
    a \emph{rule-based explanation} method e.g as in \citep{Ribeiro0G18,LORE19,LeiteKR22}
    \item a \emph{mechanistic (local) explanation} method, e.g. as provided in DAX \citep{DAX}, SpArX \citep{sparx}, and \citep{IITW2023}.  
    % \todo{add...including sparx...
    % also https://openreview.net/pdf?id=NpsVSN6o4ul and dax?};
    % \end{itemize}
    % \item $E$ is a \emph{global explanation} method, which may amount to 
    % \begin{itemize}
        \item 
    a \emph{mechanistic (global) explanation} e.g as in \citep{sparx};
% \todo{add refs, including sparx};
   \item a \emph{surrogate model} for interpreting $M$, e.g. \citep{KennyK19,tan2022surrogate,potyka2023explain_RF_with_BAF};
    \item $M$ itself, if  already an \emph{interpretable model} (e.g. a decision tree, or a set of rules).
    % \end{itemize}
\end{enumerate*}
%
% \todo{add a discussion: post-hoc vs ex-ante - if the model is interpretable (rules may be obtained from decision trees...);
% distinction can be blurred, e.g. feature attribution by LIME via surrogate model etc} 

% Although %intrinsically 
% interpretable models %(e.g. decision trees) 
% offer the advantage of clear explanations directly tied to the model's structure, they may limit performance compared to more complex models. In contrast, model-agnostic post-hoc explanation techniques (e.g. LIME or SHAP) can explain %the outperforming 
% any models, potentially at the cost of %some explanatory 
% \textit{faithfulness} to the explained model \cite{DBLP:conf/icml/DasguptaFM22}.

%\todo{are we using faithfulness and fidelity interchangeably? shall we use one term only? add a reference here? to be tidied up after we finalise the properties part}  JJ: changed to faithfulness since no fidelity elsewhere

% Naturally, for explanation methods to be useful for contestation, they need to be \emph{faithful} to $M$.\todo{add refs and discuss: is input-output faithfulness enough? do we need structural faithfulness?}
% Also, $E$ needs to satisfy other desirable properties, notably it needs to be \emph{complete} \todo{add refs, discussion}
% \todo[inline]{possibly add \emph{robust} and discuss/add refs? why are these properties useful for contestation?}

We stress that satisfying some desirable properties will make the explanation methods more useful to the contestation. Notably, \textit{faithfulness} requires that the explanations reflect the true reasoning of the explained models~\citep{DBLP:conf/aies/LakkarajuKCL19}. This will guide the interactions in contestations towards the correct direction, and allow for more effective redress. 
% Model-agnostic post-hoc explanation techniques (e.g. LIME or SHAP) can explain %the outperforming 
% any models, potentially at the cost of %some explanatory 
% faithfulness to the explained model \cite{DBLP:conf/icml/DasguptaFM22}. In contrast, 
% interpretable models %(e.g. decision trees) 
% offer the advantage of clear explanations directly tied to the model's structure (better faithfulness), they may limit predictive performance compared to more complex models. 
% 
Another important property is \textit{robustness}~\citep{Jiangsurvey}. Non-robust explanation methods could result in drastically different explanations for two users with similar inputs~\citep{DBLP:conf/aaai/LeofanteP24}, which may jeopardise their explanatory function. In contrast, robust methods~\citep{DBLP:conf/aaai/JiangL0T23,LeofanteL23,LeofanteBR23,JiangL0T24} improve the consistency and trustworthiness of explanations and may be better suited to support contestability. 

Note that, in setting (A), an explanation is not strictly necessary, e.g. a loan applicant could contest not having received a loan no matter the reason. However, if an explanation is present, the contestation can be richer, e.g., 
%if a bank customer finds out they have not received a mortgage offer because of their (low) salary, but realises, through the explanation, that their (high) bonuses have not been factored into the salary, they can contest the decision with evidence of their bonuses: here the explanation may be a feature attribution or counterfactual explanation.
if a loan applicant finds out that the reason for the loan refusal is their career-breaks,
% as provided by a feature-attribution method, 
then
they can contest the decision with evidence that they were encouraged to take career-breaks by the employer for training purposes.
%Given that, in this setting, we are interested only in contesting the model's outputs, a (possibly model-agnostic) \emph{input-output explanation} indicating how the inputs affect the outputs of the model can suffice to give users grounds for contestation and an understanding of how to support the contestation, e.g., as in the illustrative example of the bank customer,  by rectifying information held by the machine about the input or, as envisaged in \cite{wachter}, by modifying the input using information held in the (counterfactual) explanation.

Note also that, in settings (B-C), explanations are essential ``windows'' over the model, without which contestation cannot take place. 
Specifically, in the second setting, explanations are crucial and, arguably, they need to reveal the reasoning by the model for obtaining the output, e.g. as given by (faithful) mechanistic explanations. 
Indeed, input-output explanations  alone (such as those computed by feature-attribution or counterfactual explanation methods) cannot always provide grounds for contestation. 
For instance, 
consider 
the model $M$ in Example~\ref{ex:loan} and applicant $x'$
characterised again by features $\texttt{emp}$ and $\texttt{breaks}$, with the latter due to parental leaves:
an explanation including the (instances for $X\!=\!x'$ of the) first and third rule in the model empowers
the loan applicant to contest the decision 
by objecting to its bias against  people who take parental leave; this is more powerful than contesting based on the feature $\texttt{breaks}(x)$ alone in a feature-attribution explanation.

While in the first two settings (A-B) local  explanations suffice,  in the third setting 
global explanations are needed.
%Also, arguably, they need to be again mechanistic.
For illustration, a bank manager or regulator, inspecting (interpretable) model $M$ in Example~\ref{ex:loan},
may realise of a possible bias underpinning the strict definition of $continuous\_emp$ by the first rule, disregarding the possibility of career breaks %and/or parental leaves 
by applicants.

For further illustration,
consider the case of a black-box text classifier which, taken an input text, returns a classification, e.g. the sentiment of the text, or
that the text is about a certain topic. Feature-attribution explanations for the classifications (such as those provided by LIME) % or SHAP \cite{shap} or by model-specific methods such as LRP \cite{lrp} or DeefLIFT \cite{deeplift} (tailored to neural classifiers), 
can be used to pinpoint words in the input text deemed responsible for the classification, but not how the model determines the classification based on those input words -- being thus unsuitable for the second and third settings.

\textbf{Interaction for contestability.} The process of contestation needs to be supported by suitable forms of interaction between the contested ADS and the (human or ADS) contester. Interaction may be in the form of conversations in natural language, e.g. using explanations generated by Large Language Models \citep{bills2023language}, especially if the contester is a human. Alternatively, they may be in structured format, using a formal agent communication language, 
% guided by protocols 
e.g. in the spirit of FIPA \citep{posladSpecifyingProtocolsMultiagent2007}.\footnote{\url{http://www.fipa.org/}}    
In Figure~\ref{fig:overview}, we indicate with $I$ the method used by contested ADS and contester to engage in the interactions necessary for contestability.

\textbf{The viewpoint of the contester.}
So far, we have taken the viewpoint of the ADS being contested.  The contester may be a human, as in all the existing works on AI contestability, or, alternatively another ADS. In either case, 
for contestations to be acceptable, they need to be accompanied by some \emph{grounds} for contestation, as in all earlier illustrations. 
Thus, we assume that the contester is equipped with a \emph{ground generator method} $G$.
%%%%%%%%%%%%%%%%%%%%%%
% \iffalse WRONG - NEEDS ADJUSTING
% If, in the first and second setting, it is the input to be contested, this ground  may amount to evidence for the contestation (for instance, 
% %if the explanation for not granting a loan is lack of employment about the absence of a sufficiently high salary for a mortage application, a payslip showing a higher salary
% of past training or parental leaves). 
% If, in the first setting, it is the output to be contested, the ground may amount to evidence of another input with the same features but a different output.
% In the second setting....
% In the third setting....
% \fi
%%%%%%%%%%%%%%%%%%%%%%%%%%
For instance, in setting (A)
 for Example~\ref{ex:loan}, if a loan
applicant finds out that the reason for the loan refusal is their
career-breaks and they decide to contest the decision with evidence that they were encouraged to take career-breaks by the employer for
training purposes, this evidence forms the grounds for the contestation.
%\todo{add commentary for both other settings,  leveraging on previous illustrations for the loan or the text classifier} 

\textbf{Redress.} While explanations empower contestation, its complete realisation needs the ADS to have the ability to redress any issues (successfully) brought about in the contestation. Thus, we assume that the ADS is also equipped with a \emph{redress method} $R$.
For illustration, in the loan application 
Example~\ref{ex:loan}, $R$ may amount to revising the first rule by allowing for exceptions to be made when the breaks are due to training or parental leaves.  
As an additional illustration  for text classifiers, redress may result from the  a post-hoc reasoning process with the classifier's outputs and 
external knowledge encoded in argument schemes, as in \citep{lucas},
or with the classifier's outputs and
explanations therefor, e.g. as in \citep{freedman2024argumentative}. 

Note that we see contestation, and thus redress, as a post-training process, rather than model debugging during training, such as via data augmentation (e.g. as in \citep{tesoExplanatoryInteractiveMachine2019}) or regularisation (e.g. as in \citep{rossRightRightReasons2017,riegerInterpretationsAreUseful2020,shaoRightBetterReasons2021,zhangTargetedActivationPenalties2024}). Thus, our view is that contestation is, in general, different from explanation-based model debugging (e.g. as in \citep{ghai2021explainable} for tabular data, \citep{text-debug-survey} for text classification,  or \citep{popordanoska2020machine} for image classification). We, however, envisage that, in the case of models trained from data, redress may at times involve fine-tuning or retraining steps in which case the above methods, as well as methods for repairing AI models, e.g.~\citep{HenriksenLL22,AlmogK23}, could be drawn upon. 
In the illustrative case of a black-box text classifier,
an example of contestation of the full model (the third setting) is offered by FIND~\citep{FIND}: here, LRP \citep{lrp} is used to associate output neurons in the feature extractor of a text classifier with word clouds. Users can then contest the use of individual neurons by disabling them and fine-tuning the model to no longer rely on them. This allows, specifically, to decrease model bias and reliance on artifacts~\citep{FIND}. \cite{dreyerHopeSafetyUnlearning2024}, similarly, aim to expose visual concepts learned by an image classifier using concept activation vectors~\citep{kimInterpretabilityFeatureAttribution2018a} and fine-tune the model to mitigate the use of specific concepts using gradient regularisation.

\section{The Role of Computational
Argumentation% for Contestable AI
}
\label{sec:CA}

\change{In this position paper we} argue that computational argumentation (CA) is ideally suited to support AI contestability computationally. \change{CA, broadly understood as in \citep{AImagazine17,handbook}, is a branch of Knowledge Representation \& Reasoning which represents information in terms of arguments and dialectical relations (of attack and, possibly, support) between them. %Crucially, this includes  relations of support and attack, which CA analyses 
CA is equipped with semantics to reach some form of consensus regarding conclusions to be drawn. As such, CA is ideally and uniquely suited to cover all aspects of the abstract view %of contestable AI 
in Figure~\ref{fig:overview} organically.} 
% 
% More strongly, we argue that it is ideally and uniquely suited, as it can cover all aspects of the abstract view %of contestable AI 
% in Figure~\ref{fig:overview} organically.
We support this view with reference to several lines of work in the CA literature, as follows.

\textbf{CA for explanation.} CA has been widely used for XAI (see \citep{Vassiliades_21,Cyras_21} for recent overviews).
It can provide abstractions of several existing, widely used models, e.g. as in \citep{potyka2023explain_RF_with_BAF,sparx,prakken2020pagerank,fan2018generating,Cyras_19},
% [cite our work and more on ML, scheduling] 
and can itself directly serve as the basis of models \citep{rago2018argumentation,rago2020argumentation,Cocarascu_20}.
% [cite our work and more, eg on cbr]
Natural forms of explanations can be obtained from CA abstractions, e.g. dispute trees~\citep{fan2015computing,vcyras2019explanations}, defence sets~\citep{arioua2015query} and attribution scores, e.g. gradient-based argument attributions~\citep{Xiang_2023_AAE}, Shapley-based relation attributions~\citep{amgoud2017measuring,Xiang_2024_RAE}, amongst several.

\textbf{CA for redress.} CA provides methods for 
revising~\citep{SnaithR17,BaumannB15,FalappaKS09}
and repairing knowledge bases~\citep{UlbrichtB19}.
Much research has been investigated in the context of 
forgetting~\citep{BertholdR023,BaumannGR20} and enforcement~\citep{RapbergerU23,BaumannDMW21}; 
researchers investigated the effect of
expansions~\citep{Prakken23,OikarinenW11,CayrolSL10}
and changes in the knowledge base~\citep{DoutreM18,BoothKRT13,MoguillanskyRFGS13,Niskanen20}.
In addition, incomplete argumentation frameworks~\citep{BaumeisterJNNR21,AlfanoCGPT23} 
incorporate uncertainty which enables redress on a conceptual level. 

These generic lines of work in CA are useful starting points for supporting redress of models which can be abstracted argumentatively. In addition, 
reasoning with argumentation frameworks drawn from  models is the basis for forms of redress of some (machine learning) models, e.g. as in  
\citep{lucas} and \citep{freedman2024argumentative} for natural language processing, and of some (scheduling) models, as in \citep{Cyras_19}.

\textbf{CA for interaction.} Interactions between agents, often modelled as dialogues, have been shown to be effectively supported by various forms of CA, e.g. argument schemes \citep{Panisson_21} or abstract argumentation frameworks \citep{Tarle_22}, in a number of settings. CA's formal nature allows for principled desiderata of argumentative agent protocols in these settings, as defined in \citep{McBurney_02}.
These settings include: computational persuasion \citep{Fan_12_MD,Hunter_18,Calegari_21,Donadello_22}, framed as selecting the most effective arguments for changing the mind of the other agents;
information-seeking and inquiry \citep{Black_07,Fan_15_PRIMA}, where agents share information comprising arguments which are private or public; 
and other areas such as the handling of maliciousness amongst agents~\citep{Kontarinis_15}. 
Approaches to multi-agent argumentation such as these have been been shown to be useful in various real-world applications, 
e.g. regulatory compliance~\change{\citep{Raymond_22}}, recommender systems~\citep{Briguez_14,Teze_18,rago2020argumentation,Rago_21} and, more recently, interactive XAI \citep{miller-protocol,Calegari_22,guilherme}. 
One such approach which looks to have particular promise is that of \cite{Rago_23}, where \emph{argumentative exchanges} frame interactive explanation between agents as a conflict resolution problem% and are demonstrated to support the explanation of different AI models
, while accounting for humans' cognitive biases. These CA-based approaches to modelling interactions look to have promise for human interaction with Large Language Models, as in \citep{freedman2024argumentative}.

\textbf{CA and the viewpoint of the contester.} CA-based dialogue and interaction methods, e.g. as in the aforementioned \citep{Rago_23},  already have the potential to accommodate the contester's viewpoint, leveraging on  the use of CA to provide abstractions for  underpinning $G$. Indeed, e.g. in \citep{Rago_23}, contested and contester are interchangeably seen as argumentation frameworks.

\section{\change{Discussion and Future Work}}
\label{sec:conc}
In this position paper, we have proposed an abstract view of contestable AI and advocated computational argumentation (CA) as ideally positioned to support this view.

Our analysis is restricted, for simplicity and brevity, but could be naturally extended to cover broader scenarios.
For example, we have focused on two-part\change{y} scenarios where a single ADS is contested and another ADS or a human is doing the contesting; however multi-party scenarios are also possible, e.g. when, in addition to a bank customer, a regulator contests a financial institution.
Also, 
%Further, all existing discussions envisage contestability by humans towards ADSs, while ignoring  that  it may be beneficial for ADSs to be able to contest too, e.g. to critique humans or other ADSs’ biases.
in line with the existing literature on contestability,
we have assumed that only ADSs can be contested, but it may be useful to consider the possibility that 
humans may be contested by ADSs, e.g. in the spirit of~\citep{DBLP:conf/fat/000123}.
In addition to broadening our view to these and other more complex scenarios, future work will be needed to provide evidence of our claim, by building forms of contestable AI that naturally use CA.
\change{This involves connecting what we see are the main requirements of contestability (explanations, grounds for contestation, redress, and interaction) in concrete applications and scenarios.}
\change{CA has already been used to solve several individual challenges%in academic settings (with some notable exceptions
, e.g. see~\citep{BorgB20,LawrenceVR23,Cyras_19}}.
However, \change{considerable engineering} work is needed to combine solutions towards fully-fledged end-to-end contestable systems. \change{For instance, in order to support contestability with modern AI systems, we envisage that %software engineering  work will be needed 
considerable effort will be required to build large-scale argumentation solvers.}

Finally, we have advocated CA for contestability, but other KR techniques could provide useful support for some contestability aspects. Specifically, formal verification of AI models could be used to ensure the functional correctness of ADSs, e.g. as in~\citep{KouvarosLECML23}. Similarly, rigorous logic-based explainability techniques~\citep{NarodytskaSMIM19,Darwiche20} could be used to ensure the faithfulness and trustworthiness of explanations.

\section*{Acknowledgments}
%We are grateful to 
% % , as well as for useful comments from anonymous reviewers.
This research was partially supported by ERC under the EU's Horizon 2020 research and innovation programme (grant agreement No. 101020934), by J.P.
Morgan and the Royal Academy of Engineering under the Research Chairs and Senior Research Fellowships
scheme, by Imperial College through an Imperial College Research Fellowship and by UKRI through the CDT in Safe
and Trusted Artificial Intelligence (Grant No. EP/S023356/1) and the CDT in AI for Healthcare (Grant No. EP/S023283/1). Any views or opinions expressed herein are solely those of the authors. We thank Francis Rhys Ward for pointing the authors towards the literature on corrigibility, and Adam Gould, Avinash Kori and Andria Stylianou for helpful discussions.

% \clearpage

% \input{rebuttal}

%% The file kr.bst is a bibliography style file for BibTeX 0.99c
\bibliographystyle{kr}
\bibliography{main}

\end{document}